\RequirePackage{fix-cm}

\documentclass[smallextended]{svjour3}      
\usepackage{amsmath}
\usepackage[utf8]{inputenc}
\usepackage[english]{babel}
\usepackage{hyperref}
\hypersetup{
	colorlinks=true,
	linkcolor=blue,
	filecolor=blue,
	urlcolor=blue,
	citecolor=blue,
}
\usepackage{booktabs}
\usepackage[pdftex]{graphics}
\graphicspath{{figures/}}
\usepackage{longtable}
\usepackage{amsfonts}

\usepackage{epstopdf}
\usepackage[section]{placeins}
\usepackage{adjustbox}
\usepackage[a4paper, total={6.6in, 8in}]{geometry}
\usepackage{cite}
\usepackage{breqn}
\usepackage{ltcaption}
\usepackage{framed}
\usepackage{caption}
\usepackage[font={scriptsize,it}]{caption}
\usepackage{subcaption}
\usepackage{lipsum}
\usepackage{booktabs}
\usepackage{authblk}

\usepackage{chngcntr}
\counterwithin{figure}{section}
\usepackage{titlesec}
\usepackage{hyperref}
\hypersetup{
	colorlinks=true,
	linkcolor=blue,
	filecolor=blue,
	urlcolor=blue,
	citecolor=blue,
}
\smartqed  

\usepackage{graphicx}

\begin{document}
	
	\title{Development of fully intuitionistic fuzzy data envelopment analysis model with missing data: an application to Indian police sector}
	
	\author{Anjali Sonkariya  \and Awadh Pratap Singh  \and Shiv Prasad Yadav }
	
	\institute{Corresponding author: Anjali Sonkariya \at
		Department of Mathematics, Indian Institute of Technology Roorkee, Roorkee-247667, India. 
		\email{asonkariya1@ma.iitr.ac.in}       
		\and
		Awadh Pratap Singh \at
		University of Petrolium and Energy Studies, UPES, Dehradun, Dehradun-248007, India.
		\email{awadhp.singh@ddn.upes.ac.in}
		\and
	Shiv Prasad Yadav \at
	Department of Mathematics, Indian Institute of Technology Roorkee, Roorkee-247667, India.
	\email{spyorfma@gmail.com}}

	\date{Received: date / Accepted: date}

	\maketitle
	\begin{abstract}
		Data Envelopment Analysis (DEA) is a technique used to measure the efficiency of decision-making units (DMUs). In order to measure the efficiency of DMUs, the essential requirement is input-output data. Data is usually collected by humans, machines, or both. Due to human/machine errors, there are chances of having some missing values or inaccuracy such as vagueness/uncertainty/hesitation in the collected data. In this situation, it will be difficult to measure the efficiencies of DMUs accurately. To overcome these shortcomings, a method is presented that can deal with missing values and inaccuracy in the data. To measure the performance efficiencies of DMUs, an input minimization BCC (IMBCC) model in a fully intuitionistic fuzzy (IF) environment is proposed. To validate the efficacy of the proposed fully intuitionistic fuzzy input minimization BCC (FIFIMBCC) model and the technique to deal with missing values in the data, a real-life application to measure the performance efficiencies of Indian police stations is presented.
		
		\keywords{Data Envelopment Analysis \and Efficiency \and Intuitionistic fuzzy data \and Input Minimization \and Missing data.}
		
	\end{abstract}

	\section{Introduction}
	DEA is a linear programming (LP) based non-parametric technique used to measure the relative efficiencies of homogeneous DMUs. Charnes et al. \cite{charnes1978measuring} introduced the CCR DEA model for evaluating the relative efficiencies of homogeneous DMUs which consume multiple inputs to produce multiple outputs. The ratio of virtual output, the weighted sum of outputs, to the virtual input, the weighted sum of inputs, is known as the efficiency of a DMU. The maximum efficiency under consideration is called relative efficiency. It lies in (0, 1]. DMUs are called efficient if they have an efficiency score of 1 and inefficient if they have an efficiency score less than 1. Banker et al.  \cite{banker1984some} extended the CCR model to the BCC model by adding convexity constraint to the CCR model, which represents returns to scale (RTS). RTS indicates the quantitative change in outputs of a DMU resulting from a proportionate change in all inputs. RTS is of two types: (i) Constant returns to scale (CRS) and (ii) Variable returns to scale (VRS). The CCR model works under CRS, whereas the BCC model works under VRS. Also, VRS is again of two types: (a) Increasing return to scale (IRS) and (b) Decreasing return to scale (DRS). RTS is explained as follows: 
	
	If the change in the input quantity is $\alpha$, and it causes the change in the output quantity $\beta$, then there are three possibilities:\\
	(i) If $ \alpha=\beta$, then it is called CRS, \\
	(ii) If $\alpha<\beta$, then it is called IRS, \\
	(iii) If  $\alpha>\beta$, then it is called DRS.\\
	
	The conventional DEA models measure the performance efficiencies of homogeneous DMUs using crisp input-output data, but there is uncertainty in real-life problems. To deal with uncertainty, fuzzy and intuitionistic fuzzy theories have been developed. Zadeh \cite{zadeh1965fuzzy} introduced fuzzy mathematical modeling, in which a membership function is defined to express the fuzziness of data. Further, Atanassov  \cite{atanassov1986intuitionistic} introduced the intuitionistic fuzzy set (IFS) theory, an extension of fuzzy set (FS) theory which is more reliable to use in real-life problems. IFS considers membership function and non-membership grades, i.e., an element's acceptance and rejection values. Furthermore, fuzzy DEA (FDEA) \cite{sengupta1992fuzzy} and intuitionistic fuzzy DEA (IFDEA) \cite{daneshvar2011dea} models were introduced. Sengupta \cite{sengupta1992fuzzy} is the first to propose FDEA using fuzzy entropy, fuzzy regression, and the fuzzy mathematical program when the inputs, outputs, and prior information are inexact and imprecise. An integrated DEA and IF TOPSIS approach for analyzing the performance of several university departments, using IFS study was firstly proposed by Daneshvar Rouyendegh \cite{daneshvar2011dea}. \\
	In real-life applications, we have uncertainty and hesitation in the data; IFDEA models solve this problem. DEA results are data-centered. All conclusions about efficiency, benchmarks, and targets depend on the data provided. However, sometimes complete data is not available for the systems. While collecting the data for real-life problems, due to several reasons, input-output data for many DMUs is not accessible. In this case, we need to find an alternate data for those missing values. As per Kline \cite{kline1998structural} missing data can be treated in three ways: (a) they can be deleted, (b) they can be replaced with estimated scores, or (c) they can be modeled and approximated by the distribution of missing data depending on certain parameters. In the deletion method, delete all the values associated to missing data and then depict the result. But this method lose a lot of data information and results can't be relied on because data loss reduces the accuracy of results (\cite{gleason1975proposal},\cite{kim1977treatment},\cite{little2019statistical},\cite{malhotra1987analyzing}, \cite{raymond1986missing},\cite{roth1994missing}).
	In the replacement method, missing values are replaced by approximated values such as mean (average) (\cite{ford1976missing},
	\cite{hawkins1991overmodeled}
	\cite{kaufman1988application},
	\cite{little2019statistical},
	\cite{raaijmakers1999effectiveness},
	\cite{raymond1986missing}), total mean (\cite{little1988missing}
	\cite{raaijmakers1999effectiveness}), subgroup mean \cite{ford1976missing}, case mean (\cite{nie1975spss},\cite{raymond1986missing}), regression imputation (\cite{cohen1983applied}
	\cite{frane1976some},
	\cite{little1988missing},
	\cite{little2019statistical},
	\cite{raymond1987comparison}). In the distribution method, missing values are approximated by different distributions using certain parameters as maximum likelihood (\cite{ali1987missing},
	\cite{desarbo1986alternating},
	\cite{lee1990analysis}), expected maximization (\cite{azen1989estimation},
	\cite{graham1993evaluating},
	\cite{laird1988missing},
	\cite{little2019statistical},
	\cite{malhotra1987analyzing},
	\cite{ruud1991extensions}). Replacement and distribution methods give approximated results but better than deletion method. As, in this method, we don't lose the data information related to missing valued DMUs and determine results for those DMUs also which is not possible in deletion method. Further, Kao \& Liu \cite{kao2000data} replaced the missing values using fuzzy numbers (FNs). To the best of our knowledge, there is no study available in literature where the uncertainty is handled by the intuitionistic fuzzy numbers (IFNs) in presense of missing data. IFNs can express the uncertainty in more realistic manner than FNs. So, in the present study, we will evaluate the efficiencies of DMUs with missing values in IF environment.
\vspace{0.2cm}

	In this paper, we have proposed a FIFIMBCC model to solve the problem of hesitation/uncertainty/vagueness in the data and applied it to police stations in India with missing values. The rest of the paper is organized as follows: Section \ref{Preliminary} includes the basic IF concept. The proposed fully IFIMBCC model is presented in Section \ref{IFIMBCC Model}. In Section \ref{Example}, a real-life application of the proposed model in the Indian police sector is given. Finally, the conclusion is presented in Section \ref{Conclusion}. 
	
\section{Preliminary} \label{Preliminary}
	\textbf{Definition 1}  (Intuitionistic fuzzy set (IFS)) \cite{atanassov1986intuitionistic} An IFS $\tilde{P}^{I}$ is
	defined as 
	$$\tilde{P}^{I}=\left\{\left(x, \mu_{\tilde{P}^{I}}(x), \nu_{\tilde{P}^{I}}(x)\right)\right\},$$
	
	where $\mathbb{X}$ is the universe of discourse,
	$\mu_{\tilde{P}^{I}}: \mathbb{X} \rightarrow[0,1]$ is called the membership function and $\nu_{\tilde{P}^{I}}: \mathbb{X} \rightarrow[0,1]$ is called the non-membership function of $\tilde{P}^{I}$. 
	
	$\mu_{\tilde{P}^{I}}(x)$ is called the membership grade of $x \in \mathbb{X} $ being in $\tilde{P}^{I}$ and $\nu_{\tilde{P}^{I}}(x)$ is called the non-membership grade of $x \in \mathbb{X} $ being in $\tilde{P}^{I}$ with the condition that
	
	$$0 \leq \mu_{\tilde{P}^{I}}(x)+\nu_{\tilde{P}^{I}}(x) \leq 1.$$
	
	The hesitation (indeterminacy) degree of an element $x \in \mathbb{X}$ being in $\tilde{P}^{I}$ is denoted by $\pi_{\tilde{P}^{I}}(x)$ and is defined by
	
	$$\pi_{\tilde{P}^{I}}(x)=1-\mu_{\tilde{P}^{I}}(x)-\nu_{\tilde{P}^{I}}(x) \hspace{0.15cm} \forall \hspace{0.15cm} x \in \mathbb{X}, \hspace{0.2cm}  0 \leq \pi_{\tilde{P}^{I}}(x) \leq 1 .$$\\ 
	
	\textbf{Definition 2} (Convex IFS) \cite{arya2019development} 
	An IFS $\tilde{P}^I=\{(x,{\mu_{\tilde{P}^I}(x)},{\nu_{\tilde {P}^I}(x)}):x \in \mathbb{X} \}$  is convex IFS if\\
	
	$ (i) \mu_{\tilde{P}^I}(\lambda_1 x_1+\lambda_2 x_2)  \geq min(\mu_{\tilde{P}^I}(x_1),\mu_{\tilde{P}^I}(x_2)) $, $\forall x_1,x_2 \in \mathbb{X}$, where $\lambda_1+\lambda_2=1$ and $\lambda_1,\,\lambda_2 \geq 0$, i.e., $\mu_{\tilde{P}^I}$ is quasi-concave over $\mathbb{X}$.\\
	
	$ (ii) \nu_{\tilde{P}^I}(\lambda_1 x_1+\lambda_2 x_2) \leq max(\nu_{\tilde{P}^I}(x_1),\nu_{\tilde{P}^I}(x_2))$, $\forall x_1,x_2  \in \mathbb{X}$, where $\lambda_1+\lambda_2=1$ and $\lambda_1,\,\lambda_2 \geq 0$, i.e., $\nu_{\tilde{P}^I}$ is quasi-convex over $\mathbb{X}$.\\

	\textbf{Definition 3} (Intuitionistic fuzzy number (IFN))
	\cite{arya2019development} 
	A convex IFS $\tilde{P}^I=\{(x,{\mu_{\tilde{P}^I}(x)},{\nu_{\tilde {P}^I}(x)}):x \in \mathbb{R} \}$  is called an IFN if 
	$\exists$ a unique $x_0 \in \mathbb R$ such that $\mu_{\tilde{P}^I}(x_0)=1$ and $\exists$ an $x_1 \in \mathbb{R}$ such that $\nu_{\tilde{P}^I}(x_1)=1$, i.e., $\tilde{P}^I$ is normal IFS. $x_0$ is called the mean value of $\tilde{P}^I$.\\
	
\begin{figure}
		\centering
		\includegraphics[width=0.7\linewidth]{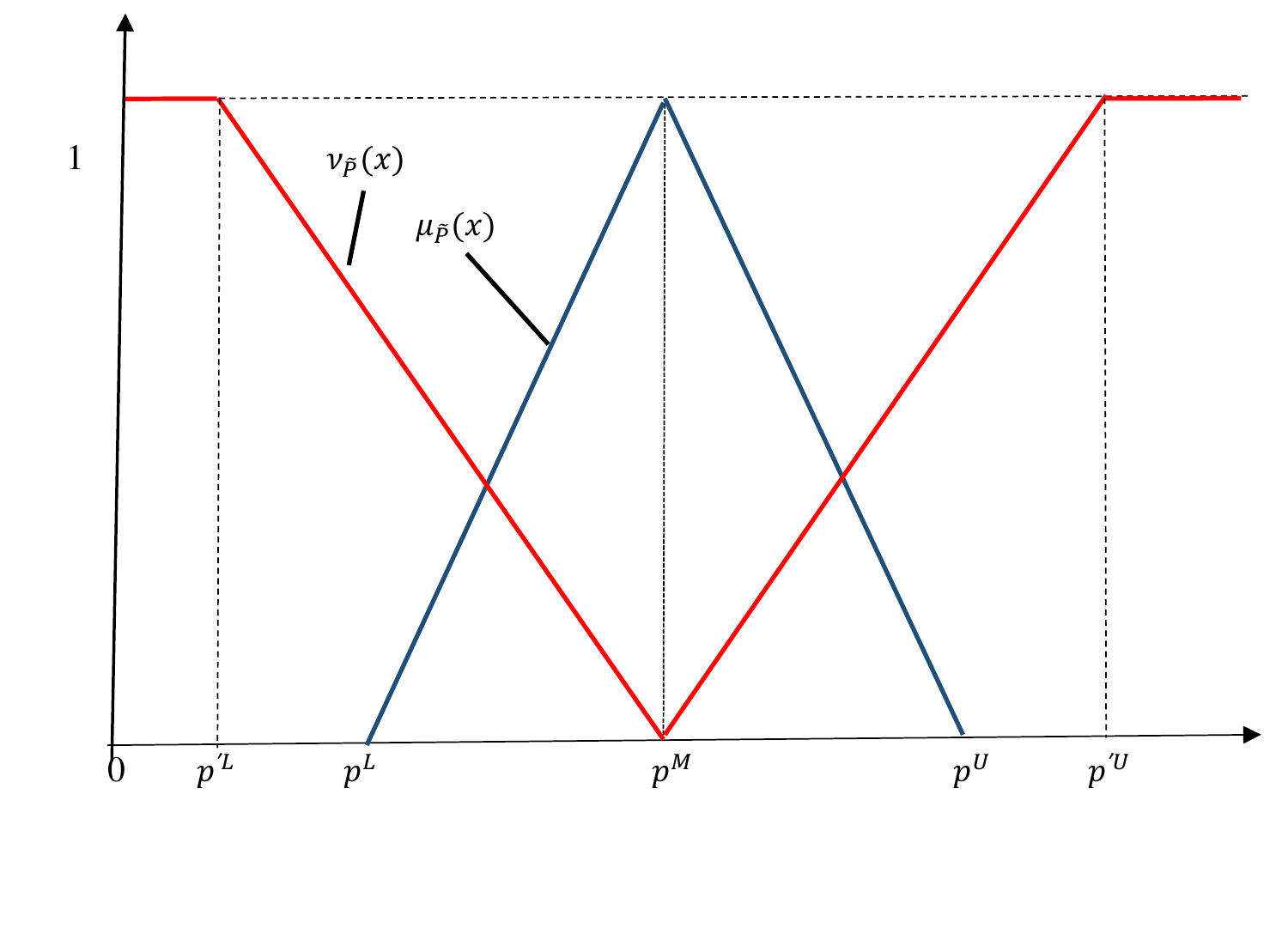}
		\caption{TIFN $\tilde P^I=(p^L,p^M,p^U;p^{\prime L},p^M,p^{\prime U})$}
		\label{TIFN}
	\end{figure}

	\textbf{Definition 4} (Triangular intuitionistic fuzzy number (TIFN)) \cite{arya2019development} The IFN\\ $\tilde {P}^I=\{(x,\mu_{\tilde P}^I(x),\nu_{\tilde P}^I(x)):\,x\in \mathbb{R} \}$ is called TIFN if\\
	\begin{minipage}[b]{0.45\textwidth}
		\[
		\mu_{\tilde{P}^I}(x)=\left\{
		\begin{array}{ll}
			\dfrac {x-p^L}{p^{M}-p^L},\, p^L < x \leq p^{M},\\
			\dfrac {p^U-x}{p^U-p^{M}}, \, p^{M} \leq x < p^U, \\
			0\,\,\,\,\,\,\,\,\,\,\,\,\,\,\,\,\,\,\,\,\,\,\,\,\,\,\,\,\,, otherwise.
		\end{array}
		\right.
		\]
	\end{minipage}	
	\hfill
	\begin{minipage}[b]{0.45\textwidth}
		\[
		\nu_{\tilde{P}^I}(x)=\left\{
		\begin{array}{ll}
			\dfrac {p^{M}-x}{p^{M}-p^{\prime L}},\, p^{\prime L} < x \leq p^{M},\\
			\dfrac {x-p^{M}}{p^{\prime U}-p^{M}}, \, p^{M} \leq x <{p^{\prime U}}, \\
			1\,\,\,\,\,\,\,\,\,\,\,\,\,\,\,\,\,\,\,\,\,\,\,\,\,\,\,\,\,, otherwise,
		\end{array}
		\right.
		\]
	\end{minipage}\\
	where $p^L,p^{M},p^U,p^{\prime L},p^{\prime U} \in \mathbb{R}$ such that  $p^{\prime L} \leq p^L \leq  p^{M} \leq p^U \leq p^{\prime U}$. TIFN is denoted by $\tilde P^I=(p^L,p^M,p^U;p^{\prime L},p^M,p^{\prime U})$. Its graphical representation is given in Figure \ref{TIFN}.
	\\

	\textbf{Definition 5} (Arithmetic operations on TIFNs) \cite{arya2019development}	Let $\tilde {P}^I= (p^L,
p^{M},p^U;p^{\prime L},p^{M},p^{\prime U})$ and $\tilde {Q}^I= (q^L,q^{M},q^U;q^{\prime L},q^{M},q^{\prime U})$ be two TIFNs. Then the arithmetic operations on TIFNs  are defined as follows:\\

(i) \textbf{Addition}:	$\tilde {P}^I\oplus \tilde {Q
}^I= (p^L+q^L,p^{M}+q^M,p^U+q^U;p^{\prime L}+q^{\prime L},p^{M}+q^M,p^{\prime L}+q^{\prime U})$,\\

(ii) \textbf{Subtraction}: $\tilde {P}^I\ominus \tilde {Q}^I= (p^L-q^U,p^{M}-q^M,p^U-q^L;p^{\prime L}-q^{\prime U},p^{M}-q^M,p^{\prime L}-q^{\prime U})$,\\

(iii) \textbf{ Multiplication}:
$ \tilde {P}^I\otimes \tilde {Q}^I \approx (p^L q^L,p^M q^M, p^U q^U;p^{\prime L} q^{\prime L},p^{M} q^M,p^{\prime L} q^{\prime U})
,\,\, \text{where}\,\, p^{\prime L},q^{\prime L} >0$.\\

(iv)	\textbf{Division}:
$$\frac{ \tilde {P}^I}{ \tilde {Q}^I} \approx \left(  \frac{p^L}{q^U},\frac{p^M}{q^M},\frac{p^U}{q^L}; \frac{p^{\prime L}}{q^{\prime U}},\frac{p^{ M}}{q^{ M}},\frac{p^{\prime U}}{q^{\prime L}}\right)\, \text{where} \,\,p^{\prime L},q^{\prime L} >0.
$$\\

	(v) \textbf{Scalar multiplication}:
	$ \text{If}\,\, \lambda \in \mathbb R,\, \text{then}$
	\[	\lambda{\tilde{P}^I}=\left\{
	\begin{array}{ll}	(\lambda p^L,\lambda p^{M},\lambda p^U;\lambda p^{\prime L},\lambda p^{M}, \lambda p^{\prime L}),\, for \,\lambda\geq 0,\\
		(\lambda p^U,\lambda p^{M},\lambda p^L;\lambda p^{\prime U},\lambda p^{M}, \lambda p^{\prime L}),\, for \,\lambda<0. \\
	\end{array}
	\right.\]
	
	\textbf{Definition 6} (Expected value of a TIFN) \cite {grzegorzewski2003distances}
	Let $\tilde P^I=(p^L,p^M,p^U;p^{\prime L},p^M,p^{\prime U})$ be a TIFN. Then its expected value is defined as\\
	$$\text{E.V.}= \frac{1}{8} \left( p^{\prime L}+ p^L+4\,p^M+p^U+p^{\prime U}\right)$$

	\section{Proposed FIFIMBCC model} \label{IFIMBCC Model} 
	Banker et al. \cite{banker1984some} proposed the BCC model with the addition of convexity constraint in the CCR model \cite{charnes1978measuring}. There are two types of BCC models in the multiplier form (i) Output Maximization Model - maximizes output quantity while using the same level of input quantity (ii) Input Minimization Model - which minimizes input quantity while producing the same level of output quantity. The conventional input minimization BCC (IM BCC) is presented below (Model 1).\\
	
	Suppose that we have $n$ DMUs and each DMU has $m$ inputs and $s$  outputs. Then, for DMU$_k, k=1,2,…,n$, the relative efficiency score $E_{k}$ is obtained from Model 1 as follows:\\
	
	\textbf{Model 1} IMBCC Model:
	$$\text{min}\,\, E_k =  \sum_{i=1}^{m} x_{ik} u_{ik} + u_{k}$$
	\hspace{5cm}	subject to
	$$\sum_{r=1}^{s} y_{rk} v_{rk} = 1,$$
	
	$$ \sum_{i=1}^{m} x_{ij} u_{ij} -\sum_{r=1}^{s} y_{rj} v_{rj} + u_{k} \geq 0 \quad \forall j=1,2,...,n,$$
	
	$$ u_{ik} \geq \varepsilon \,\, \forall \, i=1,2,...,m, \,\,
	v_{rk} \geq \varepsilon \,\, \forall \, r=1,2,...,s, \,\,\, \varepsilon>0,\,\,\text{and}\,\, u_{k} \,\, \text{is unrestricted in sign,} $$	
	
	where $x_{ij}$ is the quantity of the $i^{th}$ input of the $j^{th}$ DMU, $y_{rj}$ is the quantity of the $r^{th}$ output of the $j^{th}$ DMU, $u_{ij}$ and $v_{rj}$ are the weights corresponding to the $x_{ij}$ and $y_{rj}$ respectively and $\varepsilon$ is the non-archimedean infinitesimal.\\
	
	\textbf{Definition 7} (BCC - efficient) \cite {banker1984some} Let $E_{k}^{*}$ be the optimal value of $E_{k}$. Then DMU$_{k}$ is said to be {\bf BCC - efficient} if $E_{k}^{*}=1$; and DMU$_{k}$ is said to be {\bf BCC - inefficient} if $E_{k}^{*}<1$.\\

	Here, we extend the crisp IMBCC model to the IF environment. Many researchers have developed IFDEA models with input-output data as IFNs and weights as crisp numbers. But, in general, weights can alos be taken as IFNs. Here, in the proposed model, all data are taken as IFNs, i.e., input and output data are IFNs, along with all the weights are also IFNs. Hence the proposed FIFIMBCC Model is given as follows (Model 2): \\

	\textbf{Model 2} Proposed FFIMBCC Model:
	$$\text{min}\,\, \tilde{E}_k^I =  \sum_{i=1}^{m} \tilde{x}_{ik}^I \tilde{u}_{ik}^I + \tilde{u}_{k}^I$$
	\hspace{5cm}	subject to
	$$\sum_{r=1}^{s} \tilde{y}_{rk}^I \tilde{v}_{rk}^I = \tilde{1}^I,$$
	
	$$ \sum_{i=1}^{m} \tilde{x}_{ij}^I \tilde{u}_{ij}^I -\sum_{r=1}^{s} \tilde{y}_{rj}^I \tilde{v}_{rj}^I + \tilde{u}_{k}^I \geq \tilde{0}^I \quad \forall j=1,2,...n,$$
	
	$$ \tilde{u}_{ik}^I \geq \varepsilon \,\, \forall \, i=1,2,...m, \,\,\,
	\tilde{v}_{rk}^I \geq \varepsilon \,\, \forall \, r=1,2,...,s, \,\,\, \varepsilon>0,\,\,\text{and}\,\,\tilde{u}_{k}^I \,\, \text{is unrestricted in sign}.\,\, $$\\

	Assume that IF input  $\tilde{x}_{ij}^I$, IF output  $\tilde{y}_{rj}^I$, IF weights $\tilde{u}_{ik}^I,\tilde{v}_{rk}^I$ and $\tilde{u}_{k}^I$ are all triangular intuitionistic fuzzy numbers (TIFNs). Let 
	$$\tilde{x}_{ij}^I =(x_{i j}^{L}, x_{i j}^{M}, x_{i j}^{U} ; x_{i j}^{\prime L}, x_{i j}^{M}, x_{i j}^{\prime U}),$$
	
	$$\tilde{y}_{rj}^I = (y_{r j}^{L}, y_{r j}^{M}, y_{r j}^{U} ; y_{r j}^{\prime L}, y_{r j}^{M}, y_{r j}^{\prime U}),$$
	
	$$\tilde{u}_{ij}^I =(u_{i j}^{L}, u_{i j}^{M}, u_{i j}^{U} ; u_{i j}^{\prime L}, u_{i j}^{M}, u_{i j}^{\prime U}),$$
	
	$$\tilde{v}_{rj}^I =(v_{r j}^{L}, v_{r j}^{M}, v_{r j}^{U} ; v_{r j}^{\prime L}, v_{r j}^{M}, v_{r j}^{\prime U}),$$
	\hspace{1cm} and $$\tilde{u}_{k}^I= (u_{k}^{L}, u_{k}^{M}, u_{k}^{U} ; u_{k}^{\prime L}, u_{k}^{M}, u_{k}^{\prime U}).$$
	
	Then Model 2 is transformed into the following Model 3:
	\\
	
	\textbf{Model 3}
	$$\text{min}\,\, \tilde{E}_k^I = \left( \sum_{i=1}^{m} (x_{ik}^L,x_{ik}^M,x_{ik}^U;x_{ik}^{\prime L},x_{ik}^M,x_{ik}^{\prime U})\otimes (u_{ik}^L,u_{ik}^M,u_{ik}^U;u_{ik}^{\prime L},u_{ik}^M,u_{ik}^{\prime U})\right) \oplus (u_{k}^L,u_{k}^M,u_{k}^U;u_{k}^{\prime L},u_{k}^M,u_{k}^{\prime U}) $$
	
	subject to
	
	$$ \sum_{r=1}^{s} (y_{rk}^L,y_{rk}^M,y_{rk}^U; y_{rk}^{\prime L},y_{rk}^M,y_{rk}^{\prime U} ) \otimes (v_{rk}^L,v_{rk}^M,v_{rk}^U; v_{rk}^{\prime L},v_{rk}^M,v_{rk}^{\prime U} ) =(1,1,1;1,1,1)$$
	
	\[
	\left( \sum_{i=1}^{m} (x_{ij}^L,x_{ij}^M,x_{ij}^U;x_{ij}^{\prime L},x_{ij}^M, x_{ij}^{\prime U}) \otimes (u_{ij}^L,u_{ij}^M,u_{ij}^U;u_{ij}^{\prime L},u_{ij}^M, u_{ij}^{\prime U})\right) \ominus \left( \sum_{r=1}^{s} (y_{rj}^L,y_{rj}^M,y_{rj}^U;y_{rj}^{\prime L},y_{rj}^M, y_{rj}^{\prime U}) \right.
	\]
	
	\[
	\otimes  (v_{rj}^L,v_{rj}^M,v_{rj}^U;v_{rj}^{\prime L},v_{rj}^M, v_{rj}^{\prime U})  \bigg)  \oplus (u_{k}^L,u_{k}^M,u_{k}^U;u_{k}^{\prime L},u_{k}^M,u_{k}^{\prime U}) \geq (0,0,0;0,0,0) \quad \forall j=1,2,...,n,
	\]
	
	$$ \tilde{u}_{ik}^I \geq \varepsilon \,\, \forall \, i=1,2,...m, \,\,\,
	\tilde{v}_{rk}^I \geq \varepsilon \,\, \forall \, r=1,2,...,s, \,\,\, \varepsilon>0,\,\,\text{and}\,\,\tilde{u}_{k}^I \,\, \text{is unrestricted in sign}.\,\, $$

	Now by definition 5, using arithmetic operations of TIFNs in Model 3, we get Model 4.\\
	
	\textbf{Model 4}
	$$\text{min}\, \tilde{E}_k^I =  \left[\sum_{i=1}^{m} x_{ik}^L u_{ik}^L + u_k^L,\sum_{i=1}^{m} x_{ik}^M u_{ik}^M + u_k^M,\sum_{i=1}^{m} x_{ik}^U u_{ik}^U + u_k^U ;\sum_{i=1}^{m} x_{ik}^{\prime L} u_{ik}^{\prime L} + u_k^{\prime L},\sum_{i=1}^{m} x_{ik}^M u_{ik}^M + u_k^M,\sum_{i=1}^{m} x_{ik}^{\prime U} u_{ik}^{\prime U} + u_k^{\prime U}\right] $$
	
	subject to
	$$\left[  \sum_{r=1}^{s} y_{rk}^L v_{rk}^L, \sum_{r=1}^{s} y_{rk}^M v_{rk}^M , \sum_{r=1}^{s} y_{rk}^U v_{rk}^U \, ; \, \sum_{r=1}^{s} y_{rk}^{\prime L}  v_{rk}^{\prime L} , \sum_{r=1}^{s} y_{rk}^M v_{rk}^M , \sum_{r=1}^{s} y_{rk}^{\prime U} v_{rk}^{\prime U} \right]  \, = \, [1,1,1;1,1,1],$$

	\[ \left[ 
	\sum_{i=1}^{m} x_{ij}^L u_{ik}^L - \sum_{r=1}^{s} y_{rj}^U v_{rk}^U + u_k^L, \sum_{i=1}^{m} x_{ij}^M u_{ik}^M - \sum_{r=1}^{s} y_{rj}^M v_{rk}^M + u_k^M, \sum_{i=1}^{m} x_{ij}^U u_{ik}^U - \sum_{r=1}^{s} y_{rj}^L v_{rk}^L + u_k^U;\sum_{i=1}^{m} x_{ij}^{\prime L} u_{ik}^{\prime L} - \sum_{r=1}^{s}y_{rj}^{\prime U} v_{rk}^{\prime U} + u_k^{\prime L},
	\right.
	\]
	\[ \left. \sum_{i=1}^{m} x_{ij}^M u_{ik}^M - \sum_{r=1}^{s} y_{rj}^M v_{rk}^M + u_k^M, \sum_{i=1}^{m} x_{ij}^{\prime U} u_{ik}^{\prime U} - \sum_{r=1}^{s} y_{rj}^{\prime L} v_{rk}^{\prime L} + u_k^{\prime U} \right] \, \geq \, [0,0,0;0,0,0] , 
	\,\forall j=1,2,...,n,	\]
	
	$$ u_{ik}^{\prime L} \leq u_{ik}^L \leq u_{ik}^M \leq u_{ik}^U \leq u_{ik}^{\prime U},$$
	$$v_{rk}^{\prime L} \leq v_{rk}^L \leq v_{rk}^M \leq v_{rk}^U \leq v_{rk}^{\prime U},$$
	$$u_{k}^{\prime L} \leq u_{k}^L \leq u_{k}^M \leq u_{k}^U \leq u_{k}^{\prime U},$$
	$$ \tilde{u}_{ik}^I \geq \varepsilon \,\, \forall \, i=1,2,...m, \,\,\,
	\tilde{v}_{rk}^I \geq \varepsilon \,\, \forall \, r=1,2,...,s, \,\,\, \varepsilon>0,\,\,\text{and}\,\,\tilde{u}_{k}^I \,\, \text{is unrestricted in sign}.\,\, $$	
	
	To defuzzify Model 4, we use Definition 6 for expected value of TIFNs.
	Replacing each TIFN by expected values of the TIFNs, we get Model 5.\\
	
	\textbf{Model 5}
	$$\text{min}\,\, E_k = \frac{1}{8} \left[\sum_{i=1}^{m} x_{ik}^{\prime L} \, u_{ik}^{\prime L} + u_k^{\prime L} +  \sum_{i=1}^{m} x_{ik}^L \, u_{ik}^L + u_k^L + 4( \sum_{i=1}^{m} x_{ik}^M \, u_{ik}^M + u_k^M) + \sum_{i=1}^{m} x_{ik}^U \, u_{ik}^U + u_k^U + \sum_{i=1}^{m} x_{ik}^{\prime U} \, u_{ik}^{\prime U} + u_k^{\prime U} \right] $$
	
	subject to
	
	$$ \frac{1}{8} \left[ \sum_{r=1}^{s} y_{rk}^{\prime L}  v_{rk}^{\prime L}+\sum_{r=1}^{s} y_{rk}^L v_{rk}^L+4(\sum_{r=1}^{s} y_{rk}^M v_{rk}^M)+\sum_{r=1}^{s} y_{rk}^U v_{rk}^U+\sum_{r=1}^{s} y_{rk}^{\prime U} v_{rk}^{\prime U} \right] =1,$$
	
	\[ \frac{1}{8} \left[ (\sum_{i=1}^{m} x_{ij}^{\prime L} u_{ik}^{\prime L} - \sum_{r=1}^{s} y_{rj}^{\prime U} v_{rk}^{\prime U} + u_k^{\prime L})+ (\sum_{i=1}^{m} x_{ij}^L u_{ik}^L - \sum_{r=1}^{s} y_{rj}^U v_{rk}^U + u_k^L) + 4(\sum_{i=1}^{m} x_{ij}^M u_{ik}^M - \sum_{r=1}^{s} y_{rj}^M v_{rk}^M + u_k^M) \right.
	\]
	
	\[\left. +  (\sum_{i=1}^{m} x_{ij}^U u_{ik}^U - \sum_{r=1}^{s} y_{rj}^L v_{rk}^L + u_k^U)+ (\sum_{i=1}^{m} x_{ij}^{\prime U} u_{ik}^{\prime U} - \sum_{r=1}^{s} y_{rj}^{\prime L} v_{rk}^{\prime L} + u_k^{\prime U})\right] \geq 0,
	\quad \forall j=1,2,...,n,	\]
	
	$$ u_{ik}^{\prime L} \leq u_{ik}^L \leq u_{ik}^M \leq u_{ik}^U \leq u_{ik}^{\prime U},$$
	$$v_{rk}^{\prime L} \leq v_{rk}^L \leq v_{rk}^M \leq v_{rk}^U \leq v_{rk}^{\prime U},$$
	$$u_{k}^{\prime L} \leq u_{k}^L \leq u_{k}^M \leq u_{k}^U \leq u_{k}^{\prime U},$$	
	$$ \tilde{u}_{ik}^I \geq \varepsilon \,\, \forall \, i=1,2,...m, \,\,\,
	\tilde{v}_{rk}^I \geq \varepsilon \,\, \forall \, r=1,2,...,s, \,\,\, \varepsilon>0,\,\,\text{and}\,\,\tilde{u}_{k}^I \,\, \text{is unrestricted in sign}.\,\, $$
	
	Model 5 is the final FIFIMBCC Model to measure the efficiency of DMUs. The overall methodology to measure the efficiency of DMUs with missing data is presented in the following flowchart (\ref{figure 2}):\\
		\begin{figure}\label{figure 2}
		\centering
		\includegraphics[scale=0.8]{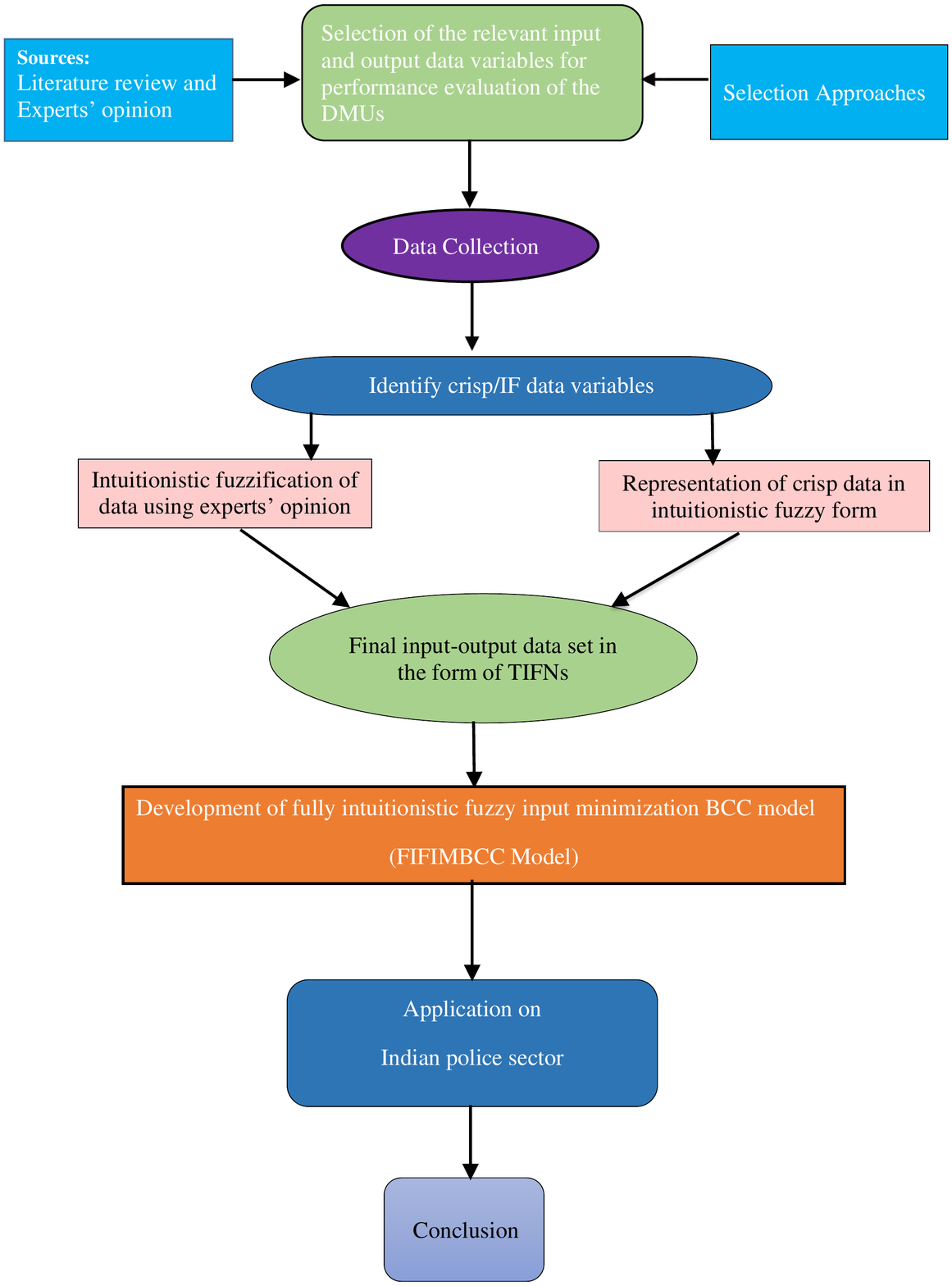}
		\caption{Efficiency of DMUs}
		\label{figure 2}
	\end{figure}

	\section{Application to Indian Police Sector} \label{Example} 
	To validate the efficacy of the proposed FIFIMBCC model in real life, an application to the Indian police sector is presented. The data is collected from the official website BUREAU OF POLICE RESEARCH AND DEVELOPMENT (BPRD), Govt. of India (As of January 1, 2018) \cite{bprd2018data}. In this study, we have considered 36 DMUs (29 States and 7 Union territories) with three inputs and three outputs. DEA results are centered on the selection of input and output data. The selection of input-output data variables is an individual choice of the decision-maker (DM). \\

	\textbf{Selection of input and output variables}\\
	Selected input-output data are given as follows:\\
	\textbf{Inputs} :\\
	I1. Total expenditure (in cr.), \\
	I2. Number of police officers,\\
	I3. Transport resources (per 100 policemen),\\
	\textbf{Outputs} :\\
	O1. Number of persons arrested,\\
	O2. Number of persons charge-sheeted,\\
	O3. Court disposal of crime cases.\\

	The collected data is crisp in nature, presented in Table \ref{tab:Table 1}. I1 of three DMUs, Jammu \& Kashmir (DMU$_{10}$), Nagaland (DMU$_{19}$), and Sikkim (DMU$_{23}$) are not available (missing). In order to apply the proposed FIFIMBCC Model, the missing values-$ X_{1,10}^I, X_{1,19}^I\, \text{and}\, X_{1,23}^I $ are converted into IFNs, particularly TIFNs. While converting crisp data values into TIFNs, membership and non-membership grades are needed. For membership grade—since the indices of DMU$_{10}$ for all input-output data aside from I1 fall within the corresponding ranges of the lowest and highest indices of all other DMUs, it is anticipated that its index of I1 will also lie within the range of the lowest index and the highest index of all other DMUs.  Thus, its smallest possible value (lower value) is set to the smallest index of this category, 21.88, and its largest possible value (upper value) is set to the largest index of this category, 15086.61. This category's median, 3105.55, is assigned as the middle-value \cite{kao2000data}. For non-membership grades—lower and upper possible values are set to be DM dependent, and the middle value is 3105.55 same as the membership function. A similar procedure is followed for DMU$_{19}$ and DMU$_{23}$. With these six values, the I1 for DMUs 10, 19, and 23 are represented by the TIFN. 
	As $ X_{1,10}^I,\, X_{1,19}^I\, \text{and}\, X_{1,23}^I $ are missing values, their corresponding TIFNs are given as follows:\\
	
	$\tilde X_{1,10}^I=(21.88,3105.55,15086.61; 18,15086.61,18000)$, \\
	
	$\tilde X_{1,19}^I=(21.88,3105.55,15086.61; 21,15086.61,15500)$
	and\\
	
	$\tilde X_{1,23}^I=(21.88,3105.55,15086.61; 20,15086.61,16800)$. \\
	
	Here, a TIFN is constructed as $(a^L,a^M,a^U;a^{\prime L},a^M,a^{\prime U})$= (min,\,median,\,max;\,\,DM,\,median,\,DM). The membership and non-membership functions of these inputs are given below:\\ 
	
	\begin{minipage}[b]{0.45\textwidth}
		\[
		\mu_{\tilde X_{1,10}^I}(x)=\mu_{\tilde X_{1,19}^I}(x)=\mu_{\tilde X_{1,23}^I}(x)=\left\{
		\begin{array}{ll}
			\vspace{0.2cm}
			\dfrac {x-21.88}{3105.55-21.88},\, 21.88 < x \leq 3105.55,\\
			\vspace{0.2cm}
			\dfrac {18000-x}{18000-3105.55}, \, 3105.55 \leq x < 18000, \\
			0\,\,\,\,\,\,\,\,\,\,\,\,\,\,\,\,\,\,\,\,\,\,\,\,\,\,\,\,\,, otherwise.
		\end{array}
		\right.
		\]
	\end{minipage}	\\
\\

	\begin{minipage}[b]{0.45\textwidth}
		\[
		\nu_{\tilde X_{1,10}^I}(x)=\left\{
		\begin{array}{ll}
			\vspace{0.2cm}
			\dfrac {3105.55-x}{3105.55-18},\, 18< x \leq 3105.55,\\
			\vspace{0.2cm}
			\dfrac {x-3105.55}{18000-3105.55}, \, 3105.55 \leq x <{18000}, \\
			1\,\,\,\,\,\,\,\,\,\,\,\,\,\,\,\,\,\,\,\,\,\,\,\,\,\,\,\,\,, otherwise.
		\end{array}
		\right.
		\]
	\end{minipage}\\
	\\

	\begin{minipage}[b]{0.45\textwidth}
		\[
		\nu_{\tilde X_{1,19}^I}(x)=\left\{
		\begin{array}{ll}
			\vspace{0.2cm}
			\dfrac {3105.55-x}{3105.55-21},\, 21< x \leq 3105.55,\\
			\vspace{0.2cm}
			\dfrac {x-3105.55}{15500-3105.55}, \, 3105.55 \leq x <{15500}, \\
			1\,\,\,\,\,\,\,\,\,\,\,\,\,\,\,\,\,\,\,\,\,\,\,\,\,\,\,\,\,, otherwise.
		\end{array}
		\right.
		\]
	\end{minipage}\\
\\
	
	\begin{minipage}[b]{0.45\textwidth}
		\[
		\nu_{\tilde X_{1,23}^I}(x)=\left\{
		\begin{array}{ll}
			\vspace{0.2cm}
			\dfrac {3105.55-x}{3105.55-20},\, 20< x \leq 3105.55,\\
			\vspace{0.2cm}
			\dfrac {x-3105.55}{16800-3105.55}, \, 3105.55 \leq x <16800, \\
			1\,\,\,\,\,\,\,\,\,\,\,\,\,\,\,\,\,\,\,\,\,\,\,\,\,\,\,\,\,, otherwise.
		\end{array}
		\right.
		\]
	\end{minipage}\\
\\

	In Kao \& Liu \cite{kao2000data} study, missing values were replaced by triangular fuzzy numbers (TFNs). In the case of the same missing $ i^{th} \text{input}/r^{th} \text{output}$ values corresponding to different DMUs, these missing input/output values should be  replaced by the same TFNs, subsequently their membership functions would also be the same.
	Here, in the present case, we have missing values in I1 for DMU$_{10}$, DMU$_{19}$, and DMU$_{23}$, so they would be assigned the same membership function. However, it is not justified to assign the same data to different DMUs as they are operating at different scales. In real-life applications, we likely have different input/output data for different DMUs. In the proposed method, missing values are replaced by TIFNs, which give the advantage of distinguishing between these missing input values for DMUs 10, 19, and 23 with the help of non-membership function values. Nevertheless, IFNs are more informative and applicable to real-life problems.\\
	
	The proposed efficiency (PE) for each DMU is calculated by applying Model 5 which is coded and solved in Lingo and the results are presented in Table \ref{tab:Table 1}. In order to calculate crisp efficiency (CE), the missing input values of DMUs 10, 19 and 23 are replaced by $X_{1,10}=5693.586,\,X_{1,19}=5381.461\,\,
	\text{and}\,\,X_{1,23}=5543.836.$ 
	The comparison of PE and CE is presented in Table \ref{tab:Table 1}. PE scores for all DMUs are close to CE scores, which validates the efficacy of the proposed model and technique to handle missing values.

	\begin{table}[htbp]
		\centering
		\caption{Indian Police Data}
		\begin{tabular}{lllllllll}
			\hline
			\textbf{DMU} & \textbf{I1} & \textbf{I2} & \textbf{I3} & \textbf{O1} & \textbf{O2} & \textbf{O3} & \textbf{PE} & \textbf{CE} \\
			\hline
			1. Andhra Pradesh & 3973.34 & 54243 & 14.48 & 156575 & 220698 & 125380 & 0.333 & 0.333 \\
			2. Arunachal Pradesh & 874.64 & 10856 & 10.56 & 2771 & 2478 & 1463 & 0.036 & 0.037 \\
			3. Assam & 4208.14 & 54535 & 8.62 & 159242 & 98525 & 44970 & 0.548 & 0.549 \\
			4. Bihar & 5398.41 & 77995 & 11.48 & 556556 & 276836 & 185498 & 1   & 1 \\
			5. Chattisgarh & 3105.55 & 59690 & 6.63 & 134135 & 132892 & 78257 & 0.849 & 0.85 \\
			6. Goa & 491.53 & 6941 & 9.78 & 3659 & 5422 & 3345 & 1   & 1 \\
			7. Gujarat & 4430.86 & 88267 & 11.83 & 456028 & 475502 & 340318 & 0.848 & 0.848 \\
			8. Haryana & 3574.89 & 44502 & 9.94 & 135703 & 113523 & 74475 & 0.41 & 0.411 \\
			9. Himachal Pradesh & 1042 & 16535 & 6.22 & 18179 & 20892 & 13899 & 1   & 1 \\
			10. Jammu \& Kasmir & *   & 77838 & 11.48 & 43210 & 42539 & 19717 & 0.084 & 0.084 \\
			11. Jharkhand & 3952.42 & 61019 & 11.69 & 57360 & 46661 & 27406 & 0.117 & 0.118 \\
			12. Karnataka & 4589.84 & 78300 & 14.85 & 270161 & 272906 & 159303 & 0.395 & 0.396 \\
			13. Kerala & 3110.03 & 44570 & 15.99 & 674544 & 724056 & 653970 & 1   & 1 \\
			14. Madhya Pradesh & 3628.44 & 93376 & 10.64 & 449590 & 555793 & 337736 & 1   & 1 \\
			15. Maharashtra & 11359.22 & 214029 & 8.26 & 535361 & 532787 & 366512 & 0.778 & 0.778 \\
			16. Manipur & 1308.56 & 24843 & 8.96 & 3185 & 1038 & 988 & 0.021 & 0.021 \\
			17. Meghalaya & 697.42 & 12691 & 9.1 & 2897 & 2544 & 1633 & 0.054 & 0.054 \\
			18. Mizoram & 478.02 & 7062 & 9.49 & 2937 & 2906 & 2461 & 1   & 1 \\
			19. Nagaland & *   & 22233 & 6.1 & 1818 & 1427 & 1181 & 1   & 1 \\
			20. Odisha & 3081.47 & 56651 & 11.48 & 112366 & 115904 & 91897 & 0.258 & 0.258 \\
			21. Punjab & 5341.09 & 82353 & 7.98 & 81907 & 73205 & 45685 & 0.255 & 0.256 \\
			22. Rajasthan & 4782.94 & 88229 & 8.29 & 254310 & 259209 & 161161 & 0.741 & 0.741 \\
			23. Sikkim & *   & 5358 & 8.27 & 1146 & 957 & 711 & 1   & 1 \\
			24. Tamil Nadu & 4660.31 & 101710 & 17.93 & 737853 & 533140 & 367791 & 1   & 1 \\
			25. Telangana & 3049.96 & 46062 & 21.73 & 88815 & 151216 & 99812 & 0.212 & 0.213 \\
			26. Tripura & 1218.57 & 23425 & 8.18 & 3421 & 4575 & 3004 & 0.038 & 0.038 \\
			27. Uttar Pradesh & 15086.61 & 285540 & 4.93 & 701359 & 826524 & 413962 & 1   & 1 \\
			28. Uttarakhand & 1618.34 & 20556 & 8.1 & 23566 & 27985 & 21995 & 0.2 & 0.198 \\
			29. West Bengal & 5384.77 & 91923 & 15.83 & 157485 & 218305 & 163036 & 0.293 & 0.294 \\
			30. A \& N Islands & 252.06 & 3958 & 16.3 & 3063 & 3387 & 2926 & 0.162 & 0.162 \\
			31. Chandigarh & 428.65 & 7748 & 9   & 5376 & 4996 & 3723 & 1   & 1 \\
			32. D\&N Haveli & 28.1 & 333 & 23.42 & 457 & 535 & 234 & 1   & 1 \\
			33. Daman \& Diu & 21.88 & 372 & 20.16 & 577 & 427 & 212 & 1   & 1 \\
			34. Delhi & 6777.02 & 74712 & 8.95 & 116114 & 81717 & 62639 & 0.32 & 0.321 \\
			35. Lakshadweep & 28.53 & 374 & 25.94 & 158 & 28  & 19  & 0.165 & 0.166 \\
			36. Puducherry & 191.27 & 2644 & 19.86 & 6156 & 5720 & 4373 & 0.178 & 0.179 \\
			\hline
			Min & 21.88 & 333 & 4.93 & 158 & 28  & 19 & &  \\
			Median & 3105.55 & 45316 & 10.25 & 69633.5 & 59933 & 36188 & & \\
			Max & 15086.61 & 285540 & 25.94 & 737853 & 826524 & 653970 & & \\
			\hline
		\end{tabular}\\
		\label{tab:Table 1}
		\hspace{-7cm} `$\ast$' indicates the value is missing from the data. 
	\end{table}

	\section{Conclusion} \label{Conclusion}
	A method to evaluate the performance efficiency of DMUs in the presence of missing data in the IF environment is developed. In real-life situations, uncertainty/vagueness/hesitation are always there. IF data models incorporate the fuzziness/vagueness as well as hesitation in the data. That is why here we developed the FIFIMBCC Model. To validate the efficacy of the proposed FIFIMBCC model in real-life, an application to the Indian police sector is presented. The collected data is crisp in nature. To apply in the present model, data is transformed into TIFNs, i.e., crisp data is intuitionistically fuzzified. Missing values  are converted into TIFNs by the proposed method.
	The results in Table \ref{tab:Table 1} depict that the proposed efficiency scores by the FIFIMBCC model are close to the crisp efficiency scores. The number of DMUs having an efficiency score of 1 in PE is the same as in CE for missing valued DMUs and other DMUs. This validates that the proposed method can handle missing values present in data impressively while measuring the performance efficiencies of DMUs. The main contribution of this paper is development of a method to handle missing data in IF enviornment. The current study initiates a research interest in the IF-DEA where uncertainty/hesitation arises from the missing data entries in DMUs. However, the model is developed using TIFNs can be further developed by using trapezoidal IFNs (TrIFNs) also.\\
		
	\textbf{Acknowledgements:} The authors are thankful to the Ministry of Education, Govt. of India, India, with grant no. MHC-01-23-200-428, 
	for financial support in pursuing this research.\\
	\\
	\textbf{Compliance with ethical standards}\\
	\\
	\textbf{Conflict of interest} Anjali Sonkariya has received research grants from The Ministry of Education, the Govt. of India, India. Shiv Prasad Yadav declares that he has no conflict of interest.\\
	\textbf{Ethical approval} This article does not contain any studies with human participants performed by any authors.\\
	\textbf{ Data Availability Statements} The datasets generated during the current study are available within the article.

	%
	%
	\bibliography{ref.bib.bib}
	\bibliographystyle{spmpsci}      
	
	%
	%
	Author, Article title, Journal, Volume, page numbers (year)
	Author, Book title, page numbers. Publisher, place (year)
	
\end{document}